\documentclass{article}

\usepackage{aaai}
\usepackage{times}
\usepackage{amsmath}
\usepackage{amssymb}
\usepackage{amsthm}
\usepackage{multirow}
\usepackage{tikz}
\usepackage{comment}

\usepackage{graphicx}
\usepackage{caption}
\usepackage{subcaption}
\usepackage{arydshln}

\newcommand{\tup}[1]{{\langle #1 \rangle}}

\newcommand{\pre}{\mathsf{pre}}     
\newcommand{\eff}{\mathsf{eff}}     
\newcommand{\cond}{\mathsf{cond}}   

\title{Generalized Planning With Procedural Domain Control Knowledge}
\author{Javier Segovia\and Sergio Jim\'enez \and Anders Jonsson\\
Dept.~Information and Communication Technologies\\
Universitat Pompeu Fabra\\
Roc Boronat 138, 08018 Barcelona, Spain\\
\{javier.segovia, sergio.jimenez, anders.jonsson\}@upf.edu
}

\begin{document}
\maketitle

\begin{abstract}
Generalized planning is the task of generating a single solution that is valid for a set of planning problems. In this paper we show how to represent and compute generalized plans using procedural Domain Control Knowledge (DCK). We define a {\it divide and conquer} approach that first generates the procedural DCK solving a set of planning problems representative of certain subtasks and then compile it as callable procedures of the overall generalized planning problem. Our procedure calling mechanism allows nested and recursive procedure calls and is implemented in PDDL so that classical planners can compute and exploit procedural DCK. Experiments show that an off-the-shelf classical planner, using procedural DCK as callable procedures, can compute generalized plans in a wide range of domains including non-trivial ones, such as sorting variable-size lists or DFS traversal of binary trees with variable size.
\end{abstract}

\section{Introduction}
\label{sec:section1}

Domain Control Knowledge (DCK) refers to an overall strategy or suggestion of how planning problems from a certain domain should be solved. There are diverse approaches for representing DCK: {\it macros}~\cite{Fikes72}, {\it control rules}~\cite{veloso:PRODIGY:JETAI1995}, {\it temporal logic formulae}~\cite{Bacchus:TLPLAN:AI99}, {\it HTNs}~\cite{Erol96}, {\it reactive policies}~\cite{fern:Gpolicies:JMLR2008,Sergio:roller:JAIR11}, {\it procedural DCK}~\cite{baier:Procedures2PDDL:ICAPS2007} or {\em finite state automata}~\cite{Geffner:FSM:AAAI10}.

\begin{figure}[hbt]
\begin{tiny}
\begin{center}
\begin{tikzpicture}[scale=.4]
  \begin{scope}
    \draw (0, 0) grid (5, 5);
    \node[anchor=center] at (3.5, 2.5) {I};
    \node[anchor=center] at (0.5, 4.5) {G4};
    \node[anchor=center] at (4.5, 4.5) {G3};
    \node[anchor=center] at (4.5, 0.5) {G2};
    \node[anchor=center] at (0.5, 0.5) {G1};
  \end{scope}
\end{tikzpicture}
\end{center}
\end{tiny}
\caption{Example navigation tasks in a $5\times 5$ grid.}
\label{fig:navigationGrid}
\end{figure}
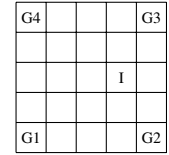

Macro-actions (i.e.~action subsequences) were among the first suggestions to speed up planning and there are several examples in the literature of computing macros~\cite{Botea:Macroff:JAIR05,coles:online-macros:JAIR207,jonsson:macros:JAIR2009}. Incorporating macros into a planning problem can help solve it faster, but even when macros are parameterized (which is not always the case), a solution involving macros may not be applicable to other problems. Consider the navigation tasks in Figure~\ref{fig:navigationGrid}, an action subsequence for reaching $G1$ starting from $I$ is no longer valid if we change the initial state.

A generalized plan is a single solution that is valid for a set of planning problems. Generalized plans are typically built with branching and repetition constructs which allows them to solve arbitrarily large problems or problems with partial observability and non-deterministic actions~\cite{Geffner:FSM:AAAI10,Infantes:FSC:ECAI2010,Levesque:GPlanning:IJCAI11,Zilberstein:Gplanning:icaps11,Giacomo:FSM:ICAPS13}. 

In this work we focus on {\it generalized plans} in the form of {\it planning programs}~\cite{Jimenez15}. Planning programs can model conditional statements and loops which allows them to represent compact solutions to individual planning tasks, as well as generalized plans. Figure~\ref{fig:program1} shows a planning program for navigating to the $(0,0)$ cell in a grid starting from any cell. Variables $x$ and $y$ represent the current position. Instructions {\small\tt dec(x)} and {\small\tt dec(y)} decrement the value of $x$ and $y$. Conditional goto instructions {\small\tt goto(0,!(x=0))} and {\small\tt goto(2,!(y=0))} jump to line $0$ when $x\neq 0$ and to line $2$ when $y\neq 0$. Finally {\small\tt end} is a marker that indicates program termination.

Since the number of possible planning programs is exponential in the number of programmable lines, our approach is to first generate procedural DCK by solving a set of planning problems representative of a certain subtask, and then compile the DCK as callable procedures in the overall generalized planning problem. We implement procedure calling using a stack modeled in PDDL so that an off-the-shelf planner can compute and exploit the DCK. Compared to previous work our contributions are:
\begin{enumerate}
\item To the best of our knowledge, ours is the first approach to compute procedural DCK over a wide range of domains and exploit it with an off-the-shelf classical planner. In previous work procedural DCK is hand-coded by human experts~\cite{baier:Procedures2PDDL:ICAPS2007} and tested in a reduced number of domains. 

\item This is the first PDDL implementation of procedural DCK with nested and recursive procedure calls. This allows an off-the-shelf classical planner to compute generalized plans for non-trivial tasks, such as sorting lists with variable size or DFS traversal of binary trees with variable size. \citeauthor{baier:Procedures2PDDL:KR08} (\citeyear{baier:Procedures2PDDL:KR08}) already formalized callable procedures but never implemented them in PDDL. \citeauthor{Jimenez15} (\citeyear{Jimenez15}) implemented callable procedures in PDDL but only for 1-level procedure calls. 

\item With respect to the original compilation~\cite{Jimenez15}, we reduce the number of actions needed to execute {\em conditional goto} instructions.
\end{enumerate}

\begin{figure}
\begin{small}
\begin{verbatim}
    0. dec(x)
    1. goto(0,!(x=0))
    2. dec(y)
    3. goto(2,!(y=0))
    4. end
\end{verbatim}
\end{small}
\caption{Generalized plan in the form of a {\it planning program} for navigating to the $(0,0)$ position in a grid.}
\label{fig:program1}
\end{figure}

\section{Background}
\label{sec:section2}

Here we introduce classical planning with conditional effects and define generalized planning as we consider it in the paper. We also review the planning programs that we use to represent generalized plans and the compilation to compute them with a classical planner.

\subsection{Classical Planning}

As is common in generalized planning, our work is based on the formulation of classical planning that includes conditional effects. This way a generalized plan can repeatedly refer to the same action, but the actual effect of the action depends on the state in which it is applied. 

We describe states and conditional effects in terms of literals. Formally, given a set of fluents $F$, a literal $l$ is a valuation of a fluent in $F$, i.e.~$l=f$ or $l=\neg f$ for some $f\in F$. A set of literals $L$ thus represents a partial assignment of values to fluents (WLOG we assume that $L$ does not assign conflicting values to any fluent). Given $L$, let $\neg L=\{\neg l:l\in L\}$ be the complement of $L$. A {\em state} $s$ is a set of literals such that $|s|=|F|$, i.e.~a total assignment of values to fluents.

A classical planning problem is a tuple $P=\tup{F,A,I,G}$, where $F$ is a set of fluents, $A$ a set of actions, $I$ an initial state and $G$ a goal condition, i.e.~a set of literals. Each action $a\in A$ has a set of literals $\pre(a)$ called the {\em precondition} and a set of conditional effects $\cond(a)$. Each conditional effect $C\rhd E\in\cond(a)$ is composed of sets of literals $C$ (the condition) and $E$ (the effect). Even though $I$ is a total assignment of values to fluents, we often describe the initial state compactly in terms of the fluents that are true in $I$.

Action $a$ is applicable in state $s$ if and only if $\pre(a)\subseteq s$, and the resulting set of {\em triggered effects} is
\[
\eff(s,a)=\bigcup_{C\rhd E\in\cond(a),C\subseteq s} E,
\]
i.e.~effects whose conditions hold in $s$. The result of applying $a$ in $s$ is a new state $\theta(s,a)=(s\setminus \neg\eff(s,a))\cup\eff(s,a)$.

A plan for $P$ is an action sequence $\pi=\tup{a_1, \ldots, a_n}$ that induces a state sequence $\tup{s_0, s_1, \ldots, s_n}$ such that $s_0=I$ and, for each $i$ such that $1\leq i\leq n$, $a_i$ is applicable in $s_{i-1}$ and generates the successor state $s_i=\theta(s_{i-1},a_i)$. The plan $\pi$ {\em solves} $P$ if and only if $G\subseteq s_n$, i.e.~if the goal condition is satisfied following the application of $\pi$ in $I$.

\subsection{Generalized Planning}

Our definition of generalized planning is based on that of \citeauthor{Giacomo:GPlanning:IJCAI11} (\citeyear{Giacomo:GPlanning:IJCAI11}), who define a generalized planning problem $\mathcal{P}=\{P_1,\ldots,P_T\}$ as a set of multiple individual planning problems that share the same observations and actions. Although actions are shared, an action may produce different results due to conditional effects. A solution to a generalized planning problem $\mathcal{P}$ is a generalized plan that solves each individual problem $P_t$, $1\leq t\leq T$. 

We restrict the above definition in two ways: 1) states are fully observable, so observations are equivalent to states; and 2) each action has the same (conditional) effects in each individual problem. As a consequence, individual problems $P_1=\tup{F,A,I_1,G_1},\ldots,P_T=\tup{F,A,I_T,G_T}$ are classical planning problems that share fluents and actions, and thus only differ in the initial state and the goal.

Early work for generalized planning followed an inductive approach, solving individual planning problems in $\mathcal{P}$ separately using a classical planner. An individual solution is then merged with the current generalized plan, and the process is repeated until the generalized plan solves the full set of problems \cite{Winner03distill:learning,Zilberstein:Gplanning:icaps11}. Another approach is compiling the generalized planning task into a conformant planning problem~\cite{Geffner:FSM:AAAI10}, i.e.~synthesizing a strong plan for a planning task with multiple possible initial states that encode the different individual planning problems under the assumption of null run-time observability. 

Generalized plans can have diverse forms that range from {\it DS-planners}~\cite{Winner03distill:learning} or {\it generalized polices}~\cite{Geffner:Gpolicies:AppliedI04}, to finite state machines (FSMs)~\cite{Geffner:FSM:AAAI10}. 
In this paper we represent and compute generalized plans using the recent formalism of {\it planning programs} ~\cite{Jimenez15} that we next proceed to define. 

\subsection{Planning Programs}

Given a classical planning problem $P=\tup{F,A,I,G}$, a planning program $\Pi$ is a numbered list of {\em instructions} such that the instruction $w$ on each line $i$ of the program is either:
\begin{enumerate}
\item A {\em sequential instruction}, i.e.~$w\in A$.
\item A {\em conditional goto instruction}, i.e.~$w=goto(i',!f)$, where $i'$ is the target program line and $f\in F$ the condition. Let $\mathcal{I}_{go}$ be the set of conditional goto instructions.
\item A {\em termination instruction} marking the end of the program.
\end{enumerate}
To execute a planning program $\Pi$ we maintain a current state $s$, initialized to $I$, and a program counter $pc$, initialized to $0$. Let $w$ be the instruction on the line indicated by $pc$. If $w\in A$, we update $s$ as a result of applying $w$ and increment $pc$. If $w=goto(i',!f)$, we set $pc$ to $i'$ if $f$ is false in the current state $s$, and increment $pc$ otherwise (as in the original paper, we jump whenever $f$ is false). Eventually, if $w$ is a termination instruction, execution ends successfully. 

Since conditional goto instructions may cause infinite loops, execution fails whenever we reach a pair of state and program counter $(s,pc)$ already visited. A planning program $\Pi$ solves a classical planning problem $P$ if the execution of $\Pi$ ends successfully and the goal condition holds in the resulting state, i.e.~$G\subseteq s$. A planning program $\Pi$ solves a generalized planning problem $\mathcal{P}=\{P_1,\ldots,P_T\}$ if it solves every classical planning problem in $\mathcal{P}$.

\subsection{Computing Programs with Classical Planning}
\citeauthor{Jimenez15}~(\citeyear{Jimenez15}) introduced a compilation that simultaneously computes a planning program and verifies that it solves a set of planning problems. Given a classical planning task $P=\tup{F,A,I,G}$ the result of the compilation is a new classical planning task $P_{n}=\tup{F_{n},A_n,I_n,G_n}$ where $n$ bounds the number of lines of the program. 

To specify $P_{n}$ we have to introduce prior notation. Let $F_{pc}=\{\mathsf{pc}_i:0\leq i\leq n\}$ be the fluents for coding the program counter and let $F_{ins}=\{\mathsf{ins}_{i,w}:0\leq i\leq n,w\in A\cup\mathcal{I}_{go}\cup\{\mathsf{nil},\mathsf{end}\}\}$ be the fluents coding the instruction $w$ on line $i$. Here, $\mathsf{nil}$ denotes a line that has not yet been programmed, while $\mathsf{end}$ denotes the end of the program. Finally, let $\mathsf{done}$ be a fluent modeling we are done programming.

For each $a\in A$, let $a_i$, $0\leq i<n$, be a classical planning action with precondition $\pre(a_i)=\pre(a)\cup\{\mathsf{pc}_i\}$ and conditional effects $\cond(a_i)=\cond(a)\cup\{\emptyset\rhd\{\neg\mathsf{pc}_i,\mathsf{pc}_{i+1}\}\}$. Likewise, for each goto instruction $goto(i',!f)\in\mathcal{I}_{go}$, let $\mathsf{go}^{i',f}_i$, $0\leq i<n$, be a classical action defined as
\begin{align*}
\pre(\mathsf{go}^{i',f}_i)=\{&\mathsf{pc}_i\},\\
\cond(\mathsf{go}^{i',f}_i)=\{&\emptyset\rhd\{\neg\mathsf{pc}_i\},\\
&\{\neg f\}\rhd\{\mathsf{pc}_{i'}\}, \{f\}\rhd\{\mathsf{pc}_{i+1}\}\}.
\end{align*}
Let $\mathsf{end}_i$, $0<i\leq n$, be a classical action defined as $\pre(\mathsf{end}_i)=\{\mathsf{pc}_i\}$ and $\cond(\mathsf{end}_i)=\{\emptyset\rhd\{\mathsf{done}\}\}$, corresponding to the termination instruction.
 
Let $w\in A\cup\mathcal{I}_{go}\cup\{\mathsf{end}\}$ be an instruction and let $w_i$ be the corresponding classical planning action that executes instruction $w$ on line $0\leq i\leq n$. Since $w$ may be executed multiple times, we define two versions: $\mathsf{P}(w_i)$, that is only applicable on an empty line $i$ and programs $w$ on that line, and $\mathsf{R}(w_i)$, that is only applicable when instruction $w$ already appears on line $i$ and repeats the execution of $w$:
\begin{align*}
\pre(\mathsf{P}(w_i))&=\pre(w_i)\cup\{\mathsf{ins}_{i,\mathsf{nil}}\},\\
\cond(\mathsf{P}(w_i))&=\{\emptyset\rhd\{\neg\mathsf{ins}_{i,\mathsf{nil}},\mathsf{ins}_{i,w}\}\},\\
\pre(\mathsf{R}(w_i))&=\pre(w_i)\cup\{\mathsf{ins}_{i,w}\},\\
\cond(\mathsf{R}(w_i))&=\cond(w_i).
\end{align*}

Now we are ready to define $P_{n}=\tup{F_{n},A_n,I_n,G_n}$:
\begin{itemize}
  \item $F_n=F\cup F_{pc}\cup F_{ins}\cup\{\mathsf{done}\}$,
  \item $\begin{aligned}[t]
&A_n=\{\mathsf{P}(a_i),\mathsf{R}(a_i):a\in A,0\leq i<n\}\\
&\cup\{\mathsf{P}(\mathsf{go}^{i',f}_i),\mathsf{R}(\mathsf{go}^{i',f}_i):goto(i',!f)\in\mathcal{I}_{go},0\leq i<n\}\\
&\cup\{\mathsf{P}(\mathsf{end}_i),\mathsf{R}(\mathsf{end}_i):0<i\leq n\},
\end{aligned}$
\item $\begin{aligned}[t]
I_n=I&\cup\{\mathsf{ins}_{i,\mathsf{nil}}:0\leq i\leq n\}\cup\{\mathsf{pc}_0\},
\end{aligned}$
\item $G_n=G\cup\{\mathsf{done}\}$. 
\end{itemize}
The compilation can be extended to a generalized planning problem $\mathcal{P}=\{P_1,\ldots,P_T\}$. After computing a planning program $\Pi$ and verifying that it solves $P_1$, simulating the end instruction resets the program counter to $0$ and the state to $I_2$, the initial state of $P_2$. To solve $P_n$, the classical plan has to simulate the execution of all planning problems in $\mathcal{P}$, thus verifying that the planning program $\Pi$ solves them all.

\subsection{Planning Programs with 1-Level Procedure Calls}
\citeauthor{Jimenez15}~(\citeyear{Jimenez15}) extended planning programs with 1-level callable procedures. Figure~\ref{fig:program2} shows a planning program with 1-level callable procedures for visiting the corners of a square grid starting from any initial position. 

\begin{figure}
\noindent
\begin{small}
\begin{tabular}{ll   |   ll}
{\tt main:} & {\tt 0.} {\tt p1} & {\tt p2:} & {\tt 0.} {\tt inc(x)}\\
 & {\tt 1.} {\tt p2} & & {\tt 1.} {\tt goto(0,!(x=n))}\\
 & {\tt 2.} {\tt p3} & & {\tt 2.} {\tt end}\\
 & {\tt 3.} {\tt p4} & & \\
 & {\tt 4.} {\tt end} & {\tt p3:} & {\tt 0.} {\tt inc(y)}\\
 & & & {\tt 1.} {\tt goto(0,!(y=n))}\\
 & & & {\tt 2.} {\tt end}\\
 & & & \\
 & & {\tt p4:} & {\tt 0.} {\tt dec(x)}\\
 & & & {\tt 1.} {\tt goto(0,!(x=0))}\\
 & & & {\tt 2.} {\tt end}
\end{tabular}
\end{small}
\caption{Planning program for visiting the 4 corners of a $n\times n$ grid starting from any initial position and using 4 auxiliary procedures ({\small\tt p1} is defined by the program in Figure~\ref{fig:program1}).}
\label{fig:program2}
\end{figure}

Planning programs with 1-level procedure calls have a new set of instructions $\mathcal{I}_{call}$, for calling auxiliary procedures (in Figure~\ref{fig:program2}, $\mathcal{I}_{call}=\{p1, p2, p3, p4\}$). Program execution proceeds as explained before except in the particular case where the instruction to execute is a calling procedure instruction. In that case the program counter is set to the first line of the called procedure and execution continues from there. Program execution always starts on the first line of the main procedure, auxiliary procedures can only be called from the main procedure, and control always returns to the main procedure when the auxiliary procedure ends. 

\citeauthor{Jimenez15}~(\citeyear{Jimenez15}) also defined a compilation to compute planning programs with 1-level procedure calls. The result of the compilation is a classical planning tasks $P_{n,b}=\tup{F_{n,b},A_{n,b},I_{n,b},G_{n,b}}$ where $b$ is a new parameter to bound the number of procedures of the planning program. $F_{n,b}$ contains the fluents of $F_n$ with the following modifications: Fluents $\mathsf{pc}_{i,j}$ and $\mathsf{ins}_{i,j,w}$ are parameterized with the associated procedure $0\leq j\leq b$ and a new fluent $\mathsf{main}$ is included to reflect that control is currently with the main procedure. Actions in $A_{n,b}$ are modified as follows:
\begin{itemize}
\item Actions $a_{i,j}$ and $\mathsf{go}^{i',f}_{i,j}$ include the associated procedure $0\leq j\leq b$, and have extra precondition $\mathsf{main}$ for $j=0$. 
\item New actions $\mathsf{call}^j_{i,0}$ implement the call to procedure $1\leq j\leq b$ on line $0\leq i<n$ of the main program, and are defined as $\pre(\mathsf{call}^j_{i,0})=\{\mathsf{pc}_{i,0},\mathsf{main}\}$ and $\cond(\mathsf{call}^j_{i,0})=\{\emptyset\rhd\{\neg\mathsf{pc}_{i,0},\neg\mathsf{main},\mathsf{pc}_{i+1,0},\mathsf{pc}_{0,j}\}\}$.
\item New actions $\mathsf{end}_{i,j}$ are defined differently for the main procedure, $j=0$, and for the auxiliary procedures $j>0$:
\begin{align*}
\pre(\mathsf{end}_{i,0})&=\{\mathsf{pc}_{i,0},\mathsf{main}\},\\
\cond(\mathsf{end}_{i,0})&=\{\emptyset\rhd\{\mathsf{done}\}\},\\
\pre(\mathsf{end}_{i,j})&=\{\mathsf{pc}_{i,j}\}, j>0,\\
\cond(\mathsf{end}_{i,j})&=\{\emptyset\rhd\{\neg\mathsf{pc}_{i,j},\mathsf{main}\}\}, j>0.
\end{align*}
\end{itemize}

\section{Nested Procedure Calls}
\label{sec:section3}
In this section we extend the 1-level procedure calls of \citeauthor{Jimenez15}~(\citeyear{Jimenez15}) to nested and recursive procedure calls. In particular, we represent a {\em call stack} and extend the semantics of {\em call} and {\em termination} instructions:
\begin{itemize}
\item  {\em Call} instructions now (1) increment the current program counter; and (2) push information onto the stack about the new procedure and its program counter.
\item  {\em Termination} instructions now pop information about the current procedure and program counter from the stack.
\end{itemize}

\subsection{The stack model}
We model a stack of finite size inspired by the compilation of {\it fault tolerant planning} into classical planning~\cite{Carmel:KfaultsPlanning:ICAPS13}. We define the stack as a tuple $\tup{K,m}$ where $K\subseteq F$ is a given set of {\it stackable fluents}, i.e., fluents that can be allocated in the stack, and $m$ is the maximum size of the stack. Implicitly our stack model defines:
\begin{itemize}
\item A set of fluents $K^m=\{f^k:f\in K,1\leq k\leq m\}$ that contains replicas of the fluents in $K$ parameterized with the stack level $k$. These fluents represent the $m$ partial states that can be stored in the stack.
\item A set of fluents $F_{top}^m=\{\mathsf{top}^k\}_{1\leq k\leq m}$ representing the top level of the stack at the current time.
\item Actions $\mathsf{push}_Q$ and $\mathsf{pop}$ are the canonical stack operations, with $\mathsf{push}_Q$ pushing a subset of stackable fluents $Q\subseteq K$ to the top level of the stack and $\mathsf{pop}$ popping any fluent in $K$ from the top level of the stack. 
\end{itemize}

\subsection{Planning Programs with Nested Procedure Calls}
Here we formalize our extension to planning programs with nested and recursive procedure calls. We add a third compilation parameter $m$ that models the stack size and hence bounds the depth of allowed nested procedure calls. The result of our extended compilation is then a classical planning task $P_{n,b}^m=\tup{F_{n,b}^m,A_{n,b}^m,I_{n,b}^m,G_{n,b}^m}$ defined as:
\begin{itemize}
\item $F_{n,b}^m=(F\setminus K)\cup K^m\cup F_{pc}^m\cup F_{ins}\cup F_{top}^m\cup\{\mathsf{done}\}$, i.e.~stackable fluents in $K\subseteq F$ are replicated and $F_{pc}^m$ contains stackable fluents indicating the line and procedure currently being executed: $F_{pc}^m=\{\mathsf{pc}_i^k:0\leq i\leq n,\linebreak1\leq k\leq m\} \cup\{\mathsf{proc}_j^k:0\leq j\leq b,1\leq k\leq m\}$.

\item $A_{n,b}^m$ parameterizes the actions in $A_{n,b}$ with the stack level $k$ and adds a precondition $\mathsf{top}^k$. Also, the actions for calling and ending procedures are redefined. For each line $i$, $0\leq i<n$, pair of procedures $j,j'$, $0\leq j,j'\leq b$, and stack level $k$, $1\leq k<m$, action $\mathsf{call}^{j',k}_{i,j}$ simulates a call to $j'$ from line $i$ of procedure $j$ on stack level $k$, and action $\mathsf{end}^{k+1}_{i,j}$ simulates the termination on line $i$ of procedure $j$ on stack level $k+1$:
\begin{align}
  \pre(\mathsf{call}^{j',k}_{i,j}) & = \{\mathsf{top}^k, \mathsf{pc}_i^k, \mathsf{proc}_j^k\},\label{eq:callact1}\\
  \cond(\mathsf{call}^{j',k}_{i,j}) & = \{\emptyset \rhd\{\neg \mathsf{pc}_i^k, \mathsf{pc}_{i+1}^k,\neg \mathsf{top}^k, \mathsf{top}^{k+1}\}\}\nonumber\\
                       & \cup\;\{\emptyset \rhd\{\mathsf{pc}_0^{k+1}, \mathsf{proc}_{j'}^{k+1}\}\},\label{eq:callact2}\\
  \pre(\mathsf{end}^{k+1}_{i,j}) & = \{\mathsf{top}^{k+1}, \mathsf{pc}_i^{k+1}, \mathsf{proc}_j^{k+1}\},\nonumber\\
  \cond(\mathsf{end}^{k+1}_{i,j}) & = \{\emptyset \rhd\{\neg \mathsf{top}^{k+1}, \neg \mathsf{pc}_i^{k+1}, \neg \mathsf{proc}_j^{k+1}\}\}\nonumber\\
 \cup\;\{&\emptyset\rhd\{\mathsf{top}^k\}\}\cup\{\emptyset\rhd\{\neg f^{k+1}:f\in K\}\}.\nonumber
\end{align}
Action $\mathsf{end}_{i,0}^1$, $0<i\leq n$, is defined as $\pre(\mathsf{end}_{i,0}^1)=\{\mathsf{top}^1,\mathsf{pc}_i^1,\mathsf{proc}_0^1\}$ and $\cond(\mathsf{end}_{i,0}^1)=\{\emptyset\rhd\{\mathsf{done}\}\}$.
\item The initial state copies stackable fluents onto stack level $1$, initializes empty program lines and sets the procedure on stack level $1$ to $0$: $I_{n,b}^m=(I\setminus K)\cup\{f^1:f\in I\cap K\}\cup\{\mathsf{ins}_{i,j,\mathsf{nil}}:0\leq i\leq n,0\leq j\leq b\}\cup\{\mathsf{top}^1,\mathsf{pc}_0^1,\mathsf{proc}_0^1\}$.
\item The goal is the same as before, i.e.~$G_{n,b}^m=G\cup\{\mathsf{done}\}$.
\end{itemize}
Action $\mathsf{call}^{j',k}_{i,j}$ is defined for any procedure pair $j,j'$; specifically the case where $j=j'$ corresponds to a recursive call. The definition of $\mathsf{call}^{j',k}_{i,j}$ included here does not copy stackable fluents and thus assumes $K=\emptyset$; in a later section we redefine call actions such that they copy stackable fluents.

\subsection{Improving the Computation of Planning Programs }
Apart from implementing nested and recursive procedure calls, we introduce another improvement to the compilation that reduces the number of actions needed to execute {\em conditional goto} instructions. In the original compilation, the number of $\mathsf{goto}^{i',f}_i$ actions is $|F|\cdot n^2$, since any fluent $f\in F$ can be a condition and there are $n^2$ combinations of line pairs $(i,i')$. Here we reduce this number to $(|F|+n)\cdot n$.

The idea is to split $\mathsf{goto}^{i',f}_i$ actions into two actions: $\mathsf{eval}^{f}_i$, that evaluates condition $f$ on line $i$, and $\mathsf{jmp}^{i'}_i$, that performs the conditional jump according to the evaluation outcome. This is inspired by assembly languages that separate comparison instructions that modify flags registers, e.g., {\tt CMP} and {\tt TEST} in the {\it x86 assembly} language, from jump instructions that update the program counter according to the flag registers, e.g., {\tt JZ} and {\tt JNZ} in {\it x86 assembly}.

To implement the split we introduce two new fluents $\mathsf{acc}$ and $\mathsf{eval}$, initially false. Fluent $\mathsf{acc}$ records the outcome of the evaluation, while $\mathsf{eval}$ indicates that the evaluation has been performed. Actions $\mathsf{eval}^{f}_i$ and $\mathsf{jmp}^{i'}_i$ are defined as
\begin{align*}
\pre(\mathsf{eval}^{f}_i)&=\{\mathsf{pc}_i,\neg\mathsf{eval}\},\\
\cond(\mathsf{eval}^{f}_i)&=\{\{f\}\rhd\{\mathsf{acc}\}\}\cup\{\emptyset\rhd\{\mathsf{eval}\}\},\\
\pre(\mathsf{jmp}^{i'}_i)&=\{\mathsf{pc}_i,\mathsf{eval}\},\\
\cond(\mathsf{jmp}^{i'}_i)&=\{\emptyset\rhd\{\neg\mathsf{pc}_i,\neg\mathsf{eval}\}\}\\
&\cup\;\{\{\neg\mathsf{acc}\}\rhd\{\mathsf{pc}_{i'}\}\}\\
&\cup\{\{\mathsf{acc}\}\rhd\{\mathsf{pc}_{i+1},\neg\mathsf{acc}\}\}.
\end{align*}
Actions for {\em programming} a conditional goto remain the same, but we introduce new actions $R(\mathsf{eval}^{f}_i)$ and $R(\mathsf{jmp}^{i'}_i)$ for repeating the execution of a goto instruction.

\section{Parameterized Procedures}
\label{sec:section4}
In many programs the number of required auxiliary procedures can be reduced parameterizing the procedures, which also decreases the total number of program lines. For example, Figure~\ref{fig:program4} shows a planning program with two parameterized procedures for visiting the corners of a grid. Compared to the program in Figure~\ref{fig:program2}, the new program is significantly more compact. 

In this section we extend nested procedure calls to parameterized procedures. We do so extending the semantics of call instructions. Now, apart from updating the program counter and pushing the new procedure onto the stack, 
{\em Call instructions} also push the parameters of the procedure onto the stack.

\begin{figure}
\noindent
\begin{small}
\begin{tabular}{l@{\hspace*{5pt}}l | l@{\hspace*{5pt}}l}
{\tt main:} & {\tt 0.} {\tt p1(x)} & {\tt p1(aux):} & {\tt 0.} {\tt dec(aux)}\\
 & {\tt 1.} {\tt p1(y)} & & {\tt 1.} {\tt goto(0,!(aux=0))} \\
 & {\tt 2.} {\tt p2(x)} & & {\tt 2.} {\tt end} \\
 & {\tt 3.} {\tt p2(y)} & & \\
 & {\tt 4.} {\tt p1(x)} & {\tt p2(aux):} & {\tt 0.} {\tt inc(aux)}\\
 & {\tt 5.} {\tt end}   & & {\tt 1.} {\tt goto(0,!(aux=n))}\\ 
 & & & {\tt 2.} {\tt end} \\
\end{tabular}
\end{small}
\caption{Planning program with parameterized procedures for visiting the four corners of a $n \times n$ grid. Procedure $p1(aux)$ decrements variable $aux$ until reaching value 0 while $p2(aux)$ increments $aux$ until reaching value $n$.}
\label{fig:program4}
\end{figure}

To explain parameter passing, we expand on the notation using variables to represent values. For a given finite domain of values $D$, let $V(D)$ be a set of variables for storing a value in $D$. Let $F_{V(D)}=\{\mathsf{assign}_{v,x}:x\in D, v\in V(D)\}$ be a set of fluents encoding assignments of type $v=x$. The key characteristic of this notation is that variables are represented as {\em objects}.

We now associate each procedure $j'$ with a parameter list $(D_1,u_1),\ldots,(D_r,u_r)$ where $D_q$, $1\leq q\leq r$, is a finite domain and $u_q\in V(D_q)$ is a corresponding (fixed) variable. The set of stackable fluents of $j'$ is $F_{V(D_1)}\cup \cdots\cup F_{V(D_r)}$. We use the action $\mathsf{call}^{j',k}_{i,j}$ defined in Equations~\eqref{eq:callact1} and~\eqref{eq:callact2} to define the actions for calling procedure $j'$. For each variable combination $v_1,\ldots,v_r\in V(D_1)\times\cdots\times V(D_r)$, we introduce a new action $\mathsf{call}^{j',k}_{i,j}(v_1,\ldots,v_r)$ defined as
\begin{align*}
  \pre&(\mathsf{call}^{j',k}_{i,j}(v_1,\ldots,v_r)) = \pre(\mathsf{call}^{j',k}_{i,j}),\\
  \cond&(\mathsf{call}^{j',k}_{i,j}(v_1,\ldots,v_r))= \cond(\mathsf{call}^{j',k}_{i,j})\\
  \cup\;\{\{&\mathsf{assign}_{v_q,x}^k\}\rhd\{\mathsf{assign}_{u_q,x}^{k+1}\}:1\leq q\leq r,x\in D_q\}.
\end{align*}
In other words, $\mathsf{call}^{j',k}_{i,j}(v_1,\ldots,v_r)$ has the effect of {\em copying} the value of each variable $v_q$, $1\leq q\leq r$, on level $k$ of the stack to the (fixed) variable $u_q$ on level $k+1$ of the stack. Since each stack level has a separate fluent set, we can reuse variables for different procedures and procedure calls (cf.~Figure~\ref{fig:program4} where $\mathsf{p1}$ and $\mathsf{p2}$ share variable $\mathsf{aux}$).



\section{Planning Programs As Control Knowledge}
\label{sec:section5}

The compilation by ~\citeauthor{Jimenez15}~(\citeyear{Jimenez15}) for generating planning programs with a classical planner does not show clear benefit from including auxiliary procedures. Indeed the configurations of the compilation that drop auxiliary procedures and increase the number of program lines of the main program were reported to achieve better, or at least equivalent, experimental results. Considering this and that the number of programs is exponential in the number of program lines, its applicability is limited to tasks solvable with planning programs of small size (they do not report planning programs of more than 6 lines including main and auxiliary procedures). 

Here we extend the applicability of planning programs to address more challenging generalized planning tasks. This is done by incorporating existing planning programs as given DCK in the form of auxiliary procedures. We now support nested procedure calls so this given DCK can represent hierarchies of planning programs.

\subsection{Exploiting Procedural Domain Control Knowledge}
In the above compilations all program lines are initially empty. However, nothing prevents us from programming some of the instructions as part of the initial state. Specifically, we consider partial planning programs such that the program lines of the main procedure are empty, but the lines of the auxiliary procedures are already programmed. For example, when generating the planning program shown in Figure~\ref{fig:program2}, the five program lines of the main procedure are initially empty but the lines of the four auxiliary procedures $p1$, $p2$, $p3$ and $p4$ appear as programmed in the initial state of the classical planning problem. 

The benefits of this approach compared to the original compilation are:
\begin{itemize}
\item The number of empty program lines is reduced, decreasing the number of possible planning programs and hence the complexity of the planning problem resulting from the compilation.
\item Each procedure solves a clearly defined subtask so the benefit of decomposing a program into procedures becomes more apparent (depending, of course, on how the procedure is obtained).
\end{itemize}
Evidently, the benefit of including a given procedure is contingent on how much the procedure contributes to solving the overall problem. The question, then, is how to generate useful procedures. One option is for a domain expert to hand-craft auxiliary procedures \cite{baier:Procedures2PDDL:ICAPS2007}, and this might be the best choice if such control knowledge is readily available. 

\subsection{Automatic Generation of Procedural DCK}
Here we show how to take another approach to generating useful procedural DCK, which is to automatically compute the auxiliary procedures. Since each procedure is in effect a program of its own, we can use the same compilation as in \citeauthor{Jimenez15}~(\citeyear{Jimenez15}) to compute this program from examples (albeit without auxiliary procedures). In the navigation example from Figure~\ref{fig:navigationGrid}, to compute the auxiliary procedure $p1$, i.e., a program for navigating to position $(0,0)$, we simply define a series of planning problems with different initial states whose goal condition is to be at position $(0,0)$. Similarly, we can define suitable planning problems to compute the remaining auxiliary procedures in Figure~\ref{fig:program2}.

We assume the existence of a specific decomposition of the overall problem into a set of subtasks, and appropriately extend the definition of a generalized planning problem. Formally, a generalized planning problem with subtasks is a tuple $\mathcal{P}_{sub}=\langle\mathcal{P},\mathcal{P}^1,\ldots,\mathcal{P}^b\rangle$, where $\mathcal{P}=\{P_1,\ldots,P_T\}$ is the generalized planning problem that we ultimately want to solve. For each $j$, $1\leq j\leq b$, $\mathcal{P}^j=\{P_1^j,\ldots,P_T^j\}$ is also a generalized planning problem whose purpose is to compute the $j$-th auxiliary procedure that corresponds to the $j$-th subtask.

Given a generalized planning problem with subtasks $\mathcal{P}_{sub}$, for each $j$, $1\leq j\leq b$, we separately compute a planning program $\Pi^j$ that solves the generalized planning problem $\mathcal{P}^j$. We then compile a partial planning program with $b$ auxiliary procedures corresponding to the planning programs $\Pi^1,\ldots,\Pi^b$, but whose main procedure has empty program lines. Finally, we compute a solution to the compiled planning problem, corresponding to a generalized plan in the form of a planning program with procedures that solves the generalized planning problem $\mathcal{P}$. This process can be iterated generating a hierarchy of planning programs. 

\begin{figure*}[hbt]
\begin{center}
  \begin{subfigure}[b]{0.23\textwidth}
		\centering
        \includegraphics[width=0.9\textwidth]{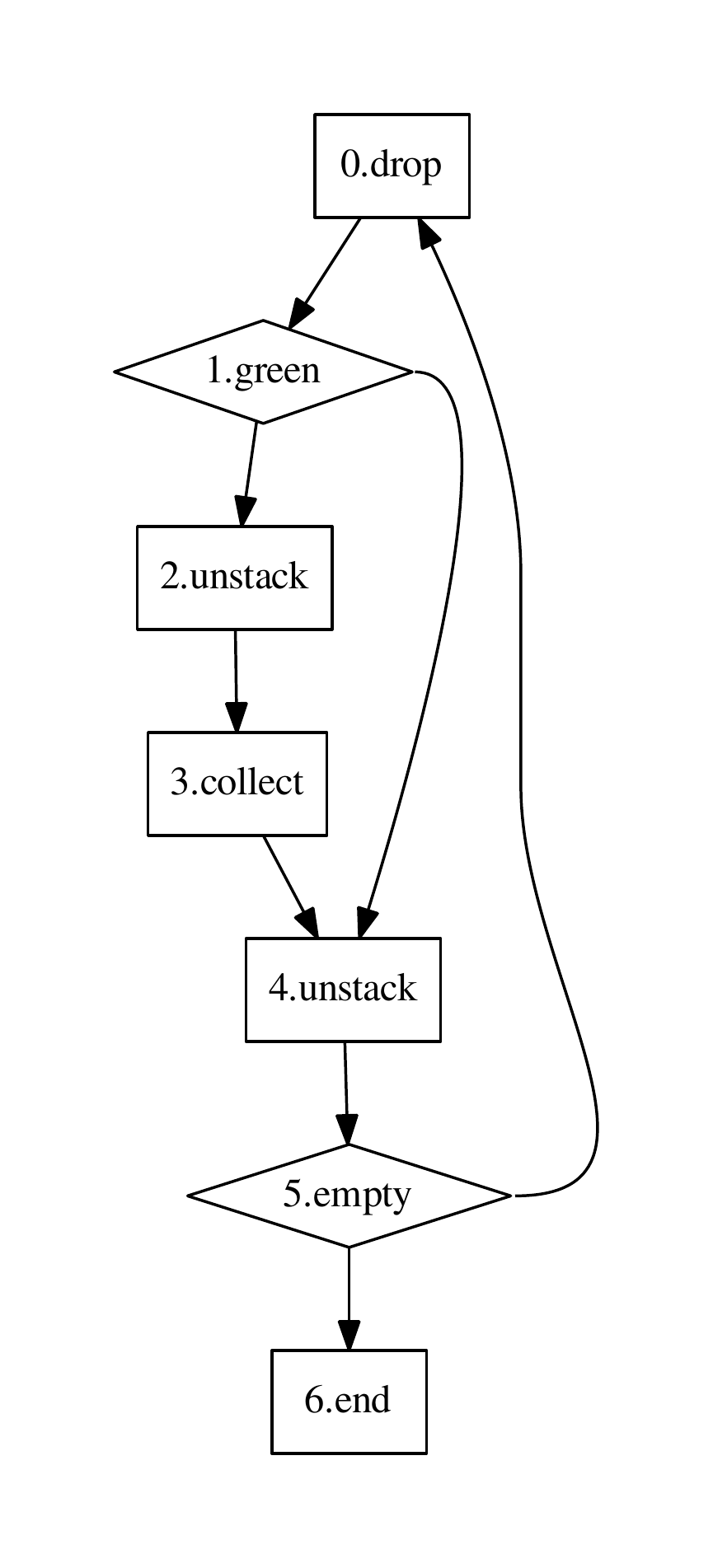}
        \caption{{\bf Blocks}}
        \label{fig:prog-blocks}
    \end{subfigure}
    \begin{subfigure}[b]{0.44\textwidth}
		\centering
        \includegraphics[width=0.55\textwidth]{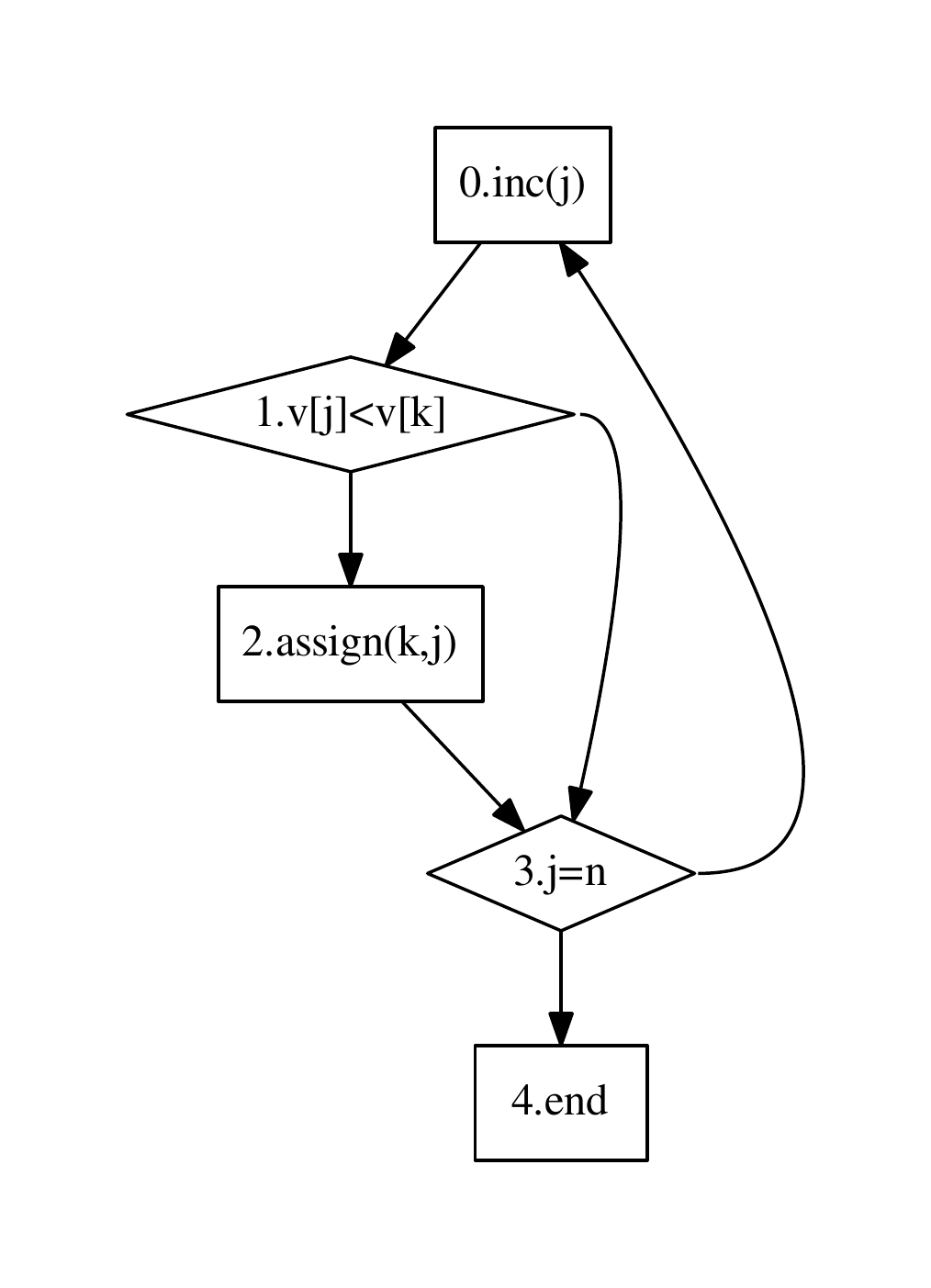}
        \includegraphics[width=0.4\textwidth]{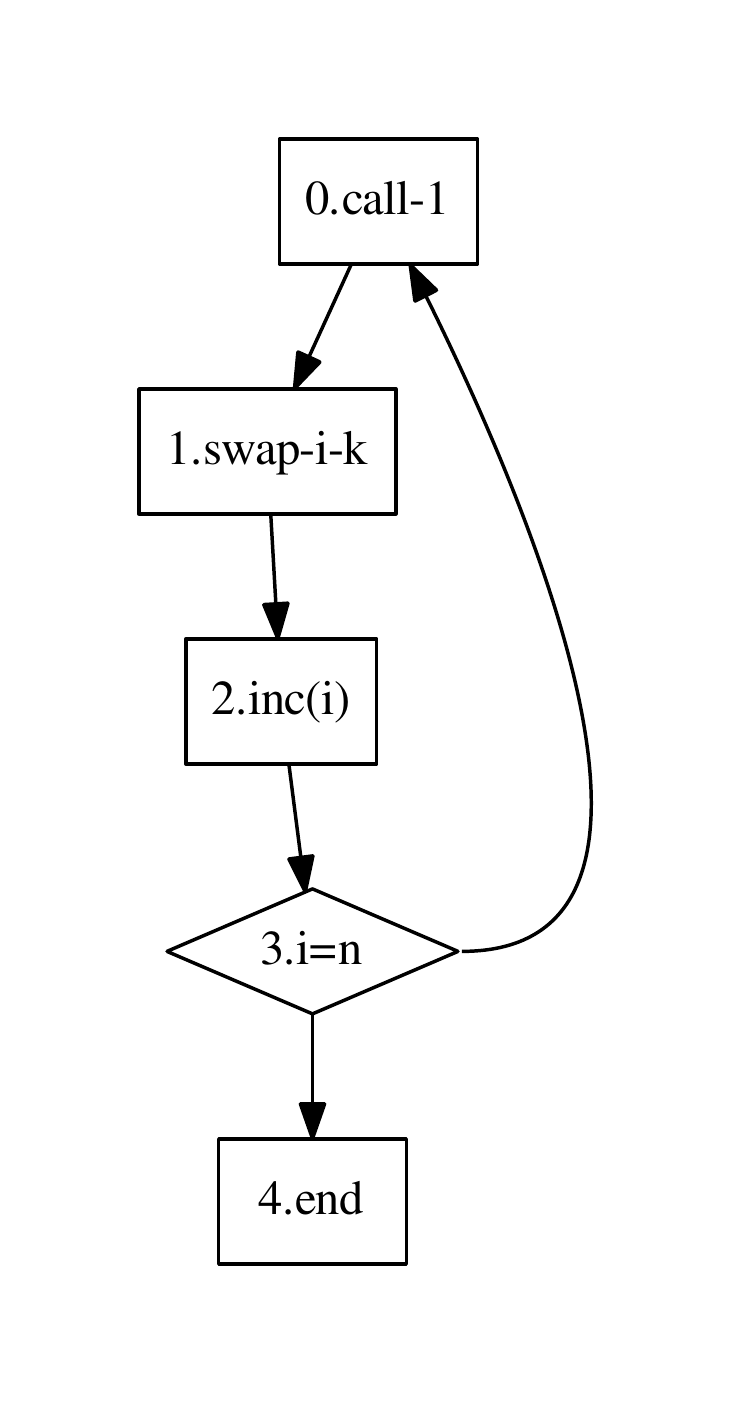}
        \caption{{\bf Sorting} (procedure 1 and main)}
        \label{fig:sorting0}
    \end{subfigure}
    \begin{subfigure}[b]{0.32\textwidth}
		\centering
        \includegraphics[width=0.92\textwidth]{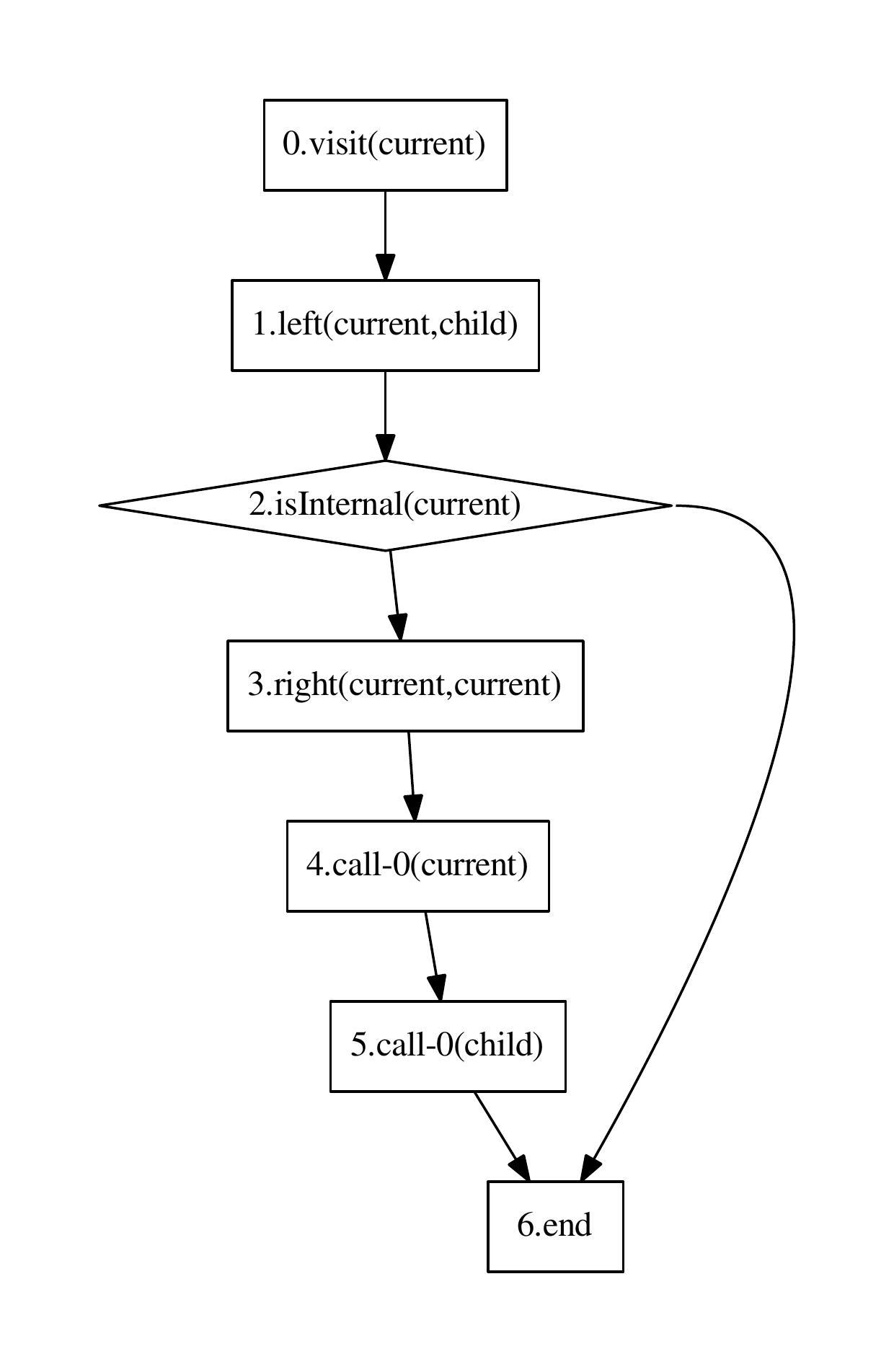}
        \caption{{\bf Tree} (recursive calls at lines 4 and 5)}
        \label{fig:prog-tree}
    \end{subfigure}
\end{center}
\caption{Flow diagrams of example planning programs obtained for selected domains.}
\label{fig:programs}
\end{figure*}

\section{Evaluation}
\label{sec:section6}

In all experiments, we run the classical planner Fast Downward~\cite{Helmert:FD:JAIR06} with the {\sc Lama-2011} setting~\cite{richter:lama:JAIR2010} on a processor Intel Core i5 3.10GHz x 4 with a 4GB memory bound and time limit of 3600s. 

We evaluate our approach in Blocks, Gripper, Hall-A and Visual-Marker from~\citeauthor{Geffner:FSM:AAAI10}(~\citeyear{Geffner:FSM:AAAI10}). In {\bf Blocks} the problem is to unstack blocks from a tower until a green block is found. In {\bf Gripper} a robot has to move balls from one room to another. In {\bf Hall-A} the problem is to visit the four corners of a grid as in Figure~\ref{fig:navigationGrid}. In {\bf Visual-Marker} the problem is moving a marker from the bottom-left corner of a grid to a cell with a green block. We also add the {\bf Grid} domain, where an agent in an arbitrary cell of a grid must visit another arbitrary cell and {\bf Visit-All}, where an agent in the bottom-left corner of a grid must visit all the grid cells.

In these domains we model integer variables $v$ with fluents of type $v=a$, and define actions {\small\tt inc(v)} and {\small\tt dec(v)} with conditional effects to increment or decrement $v$. For instance, in Grid and Visit-All variables $x$ and $y$ model the position of the agent, and $w$ and $h$ model the width and height of the grid (allowing then variable grid sizes). In Grid we also model the goal cell using variables $x_G$ and $y_G$. To obtain a general program, Grid models derived fluents $x=x_{G}$ and $y=y_{G}$ that reflect when the agent is at a goal position, and Visit-All includes $x=w$ and $y=h$ that reflect when the agent is at the last row or column. Both domains include a {\small\tt visit} action that marks the current cell as visited. 

A benefit of computing generalize plans in the form of programs is that we can naturally model programming tasks: 
\begin{itemize}
\item {\bf Summatory} models computing the summatory of a natural number $n$, $y=1+\cdots+n$, using integer variables $n$ and $y$. The initial state of each test is $(n=m,y=0)$ for some $m$ and the goal is $y=\sum_{x=1}^m x$. {\bf Fibonacci} models computing the $n^{th}$ term of the {\it Fibonacci} sequence, $F_n=F_{n-1}+F_{n-2}$, using integer variables $n$, $F_n$, $F_{n-1}$ and $F_{n-2}$. The initial state is $(n=m,F_n=1,F_{n-1}=0,F_{n-2}=0)$ for some $m$ and the goal is $F_n=F_m$.
\item {\bf Reverse} models the task of reversing a list of numbers with variable list size $n$ using integer variables $i$ and $j$ (representing iterators) as well as $n$ and $v_r$, $1\leq r\leq n$. The domain includes action {\small\tt swap-i-j} that swaps the values of $v_i$ and $v_j$. The initial state is given by $(i=1,j=n)$ and arbitrary $n$ and $v_r$, $1\leq r\leq n$, and the goal is to reverse the list represented by $v_r$, $1\leq r\leq n$. {\bf Sorting} models the task of ordering a list of numbers with variable list size $n$ using integer variables $i$, $j$ and $k$ (representing iterators) as well as $n$ and $v_r$, $1\leq r\leq n$. The domain includes action {\small\tt swap-i-k}, actions {\small\tt inc(i)} and {\small\tt inc(j)}, and an action that assigns the resulting value of $j$ to $k$. The initial state is given by $i=j=k=1$ and arbitrary $n$ and $v_r$, $1\leq r\leq n$, and the goal is to sort the list represented by $v_r$, $1\leq r\leq n$.
\item The {\bf List} domain iterates over the nodes of a linked list. We define a finite domain $D$ that includes all list nodes, and variables that point to nodes. Fluent {\small\tt is-tail} indicates the tail node, action {\small\tt next} accesses the next node and action {\small\tt visit} marks nodes as visited. The {\bf Tree} domain DFS visits all nodes of a binary tree. The domain $D$ includes tree nodes, and fluents {\small\tt is-internal} indicate internal nodes. Two actions {\small\tt left} and {\small\tt right} access the left and right child of a node, respectively, and we use variables {\small\tt current} and {\small\tt child}. The {\small\tt visit} action marks nodes as visited.
\end{itemize}

For programming tasks, given two integer variables $v$ and $w$ we add action {\small\tt assign(v,w)} to assign the value of $w$ to $v$, and {\small\tt add(v,w)}, adding the value of $w$ to the current value of $v$. We make minimal assumptions about conditions in goto instructions. Given a generalized planning problem $\mathcal{P}=\{P_1=\langle F,A,I_1,G_1\rangle,\ldots,P_T=\langle F,A,I_T,G_T\rangle\}$, any fluent in $F$ can be a condition. Moreover, for selected pairs of integer variables $v$ and $w$, we add derived fluents $(v=w)$, as well as $(v<w)$ for Reverse and Sorting.

Figure~\ref{fig:programs} shows the flow diagrams of the planning programs obtained for several evaluation domains and Table 1 summarizes the obtained experimental results. For each domain we report: number of {\bf procedures} used to solve the generalized planning problem (including the main procedure) and kind of {\bf solution} (or error) obtained. For each procedure (starting with the main procedure): {\bf program lines}, number of {\bf instances} used, planning {\bf time} (the sum of all the individual times too) and {\bf plan length}. When the solution kind is {\bf One Procedure} (OP) only a main procedure is needed to obtain a generalized plan. This corresponds to not using DCK and is the case of Blocks, Gripper, Summatory, Fibonacci and Reverse, where a general program able to solve any instance of the domain was generated without DCK. In List and Tree only 1 procedure was used but with {\bf Recursivity} (R) and {\bf Recursivity with Parameters} (RP) respectively. {\bf Nested Procedures} (NP) indicates that the number of procedures is greater than 1 so auxiliary procedures are used as DCK. In some cases, where procedural DCK pays off, the generalized planning problem could not be solved without DCK in the given memory bound, e.g., Hall-A, Grid, Sorting, Visit-All and Visual-Marker. In these domains procedural DCK was automatically generated as generalized plans for subtasks and then inserted as procedures in the overall generalized planning problem. Other domains such as Blocks, Fibonacci and Gripper can be solved without DCK, even though using DCK is better, in terms of time for the first two, and plan length for the latter.

We briefly describe the obtained auxiliary procedures. In {\bf Hall-A}, each procedure solves the task of visiting one of the 4 corners. In {\bf Grid}, one procedure solves the task of reaching the target column ($x$), while the other reaches the target row ($y$). In {\bf Visit-All}, one procedure visits all cells in a grid with a single row from left to right, while the other returns the agent to the starting (i.e.~leftmost) column. In {\bf Visual Marker}, similar to {\bf Visit-All}, one procedure moves the marker right until a green block is found (or the end of the grid is reached) and the other returns the marker back to the first column. In {\bf Fibonacci} the procedure performs one update of the Fibonacci sequence. In {\bf Sorting}, the procedure solves the generalized planning problem of selecting the smallest element in a partial list; hence the generalized plan corresponds to selection sort. The {\bf Blocks} domain has a procedure to find the green block, and {\bf Gripper} has two procedures, the first one collects two balls and moves to the other room and the second one drops the carried balls and moves to the other room. 

The compilation of \citeauthor{Jimenez15}~(\citeyear{Jimenez15}) cannot generate recursive solutions, i.e.~of types R and RP. Besides it fails to generate solutions that require a significant number of program lines (Grid, Hall-A, Sorting, Visit-all and Visual-Marker). The state-of-the-art in generalized planning is the automatic derivation of FSMs. This approach is able to solve the Hall-A, Blocks, Gripper and Visual-Marker~\cite{Geffner:FSM:AAAI10,Giacomo:FSM:ICAPS13} domains as well as Visit-All and Sorting~\cite{Zilberstein:Gplanning:icaps11}. The previous solution for Sorting assumed fixed list size and an existing action for inserting an element in a sorted list, while we allow variable list size and automatically compute the procedure for selecting the smallest element from a sublist. Remarkably, \citeauthor{Giacomo:FSM:ICAPS13}~(\citeyear{Giacomo:FSM:ICAPS13}) compute solutions to the FSM domains in a fraction of a second. FSMs can represent generalized plans more compactly than planning programs, reducing the search space, and the authors also hand-pick the conditions for branching while we allow any fluent to appear as a condition. However, there is currently no mechanism that enables procedure calls for FSMs, preventing FSMs from representing solutions to the {\bf Tree} domain.

\begin{table*}[hbt!]
\centering
\begin{footnotesize}
\begin{tabular}{l@{\hspace*{5pt}}r@{\hspace*{5pt}}r@{\hspace*{5pt}}r@{\hspace*{5pt}}r@{\hspace*{5pt}}r@{\hspace*{5pt}}r@{\hspace*{5pt}}r}
 \textbf{Domain} & \textbf{Procedures} & \textbf{Solution} & \textbf{Lines} & \textbf{Instances} & \textbf{Time(s)} & \textbf{Total time (s)} & \textbf{Plan length} \\\hline
Blocks			&	2			&	NP			&	4,3			&	5,5				&	2,2				&	4&	46,61\\
Blocks			&	1			&	OP			&	6			&	5				&	85				&	85				&	73 \\\hdashline[1pt/5pt]
Fibonacci		&	2			&	NP			&	3,3			&	2,5				&	2,177			&	179	&12,129 \\
Fibonacci		&	1			&	OP			&	5			&	2				&	3570			&	3570&56 \\\hdashline[1pt/5pt]
Grid			&	3			&	NP			&	5,5,2		&	2,2,4			&	611,631,2		&	1244			&43,43,213\\
Grid			&	1			&	ME			&	10			&	5				&	-				&	-&	-\\\hdashline[1pt/5pt]
Gripper			&	3			&	NP			&	3,3,3		&	2,2,2			&	1,1,2			&	4			&12,12,54\\
Gripper			&	1			&	OP			&	4			&	2				&	1				&	1&	77 \\\hdashline[1pt/5pt]
Hall-A			&	5			&	NP			&	5,5,5,3,5	&	2,2,2,2,2		&	3,7,3,1,4		&	18			&44,40,40,73,155\\
Hall-A			&	1			&	ME			&	14			&	2				&	-				&	-&	- \\\hdashline[1pt/5pt]
List			&	1			&	R			&	5			&	6				&	5				&	5&	120\\\hdashline[1pt/5pt]
Reverse			&	1			&	OP			&	4			&	2				&	22				&	22				&	38\\\hdashline[1pt/5pt]
Sorting			&	2			&	NP			&	4,4			&	4,3				&	14,16			&	30	&73,188 \\
Sorting			&	1			&	ME			&	7			&	4				&	-				&	-&	-\\\hdashline[1pt/5pt]
Summatory		&	1			&	OP			&	3			&	2				&	1				&	1&	26 \\\hdashline[1pt/5pt]
Tree/DFS		&	1			&	RP			&	6			&	1				&	165				&	165				&	51 \\\hdashline[1pt/5pt]
Visitall		&	3			&	NP			&	4,2,4		&	2,2,2			&	1,1,5			&	7			&47,18,238 \\
Visitall		&	1			&	ME			&	7			&	2				&	-				&	-&	- \\\hdashline[1pt/5pt]
Visual-Marker	&	3			&	NP			&	4,2,4		&	4,2,5			&	2,1,10			&	13				&82,18,205 \\
Visual-Marker	&	1			&	ME			&	8			&	2				&	-				&	-	&-
\end{tabular}
\end{footnotesize}
\caption{Procedures (1: only main; $>$1: DCK as auxiliary procedures were used), solution kind (NP=Nested Procedures, OP=One Procedure, R=Recursivity, RP=Recursivity with Parameters, ME=Memory error) and per each procedure: program lines, instances used to generate each procedure, planning time and plan length.}
\end{table*}

\section{Related Work}
\label{sec:section7}

Procedural DCK in the form of Golog-like programs ~\cite{baier:Procedures2PDDL:ICAPS2007,baier:Procedures2PDDL:KR08} include conditionals and loops as well as non-deterministic action choices that constrain the search for a solution plan. Nevertheless they are not proper generalized plans since it is still necessary to apply a planner to solve each individual planning problem. In addition they are hand-crafted and do not implement in PDDL mechanisms for procedure calling. 

Planning programs~\cite{Jimenez15} can model generalized plans and be automatically computed from test cases but allow only one level of procedure calls which makes recursion unfeasible. Moreover the original compilation does not show benefits from including auxiliary procedures so it could only address generalized planning tasks that can be solved with small planning programs. 

 Previous work on computing FSMs~\cite{Geffner:FSM:AAAI10} also uses a compilation that interleaves programming a FSM with verifying that it satisfies a set of test cases. The generation of a FSM is however different since it starts from a partially observable planning model and uses a conformant to classical planning compilation~\cite{palacios-conformant-JAIR09}.  FSMs can be understood as a way of representing and computing procedural DCK that (1) does not implement procedure calls and recursion; (2) does not reuse FSMs for similar tasks. Another difference is that our {\em conditional goto} instructions can branch on any fluent without the need to predefine a subset of observations, although we would also benefit from restricting the number of branch conditions.


\section{Conclusion}
\label{sec:section8}
In this paper we introduce a novel approach to computing and exploiting procedural DCK to solve generalized planning tasks. Our approach compiles procedural DCK into callable procedures that allow an off-the-shelf planner to solve challenging generalized planning tasks such as sorting lists with variable list size or DFS traversal of a binary tree with variable size. As far as we know this is the first approach to compute procedural DCK over a wide range of planning tasks using an off-the-shelf planner. Ours is also the first PDDL implementation of procedural DCK that handles nested procedure calls and allows representing and computing recursive solutions with an off-the-shelf planner.

In experiments, the resulting generalized plan solves all instances of a given domain. In general, however, there are no such guarantees; an extension would be to define test instances and verify that a generalized plan solves all of them. We have not formally characterized the type of plans that we can represent but intuitively, single programs (or FSMs) correspond to regular expressions, and our additions make it possible to represent plans more compactly. Some planning domains require extra high-level state features, e.g., "well-placed", "above" or "good-tower" concepts in Blocksworld~\cite{Geffner:Gpolicies:AppliedI04}, for defining compact generalized plans. It remains as future work how to automatically generate such high-level state features.

The strongest assumption of our approach is the existence of a subtask decomposition of the overall generalized planning problems, necessary to compute the DCK in the form of procedures. An interesting research direction would be to automatically discover these decompositions. This would make it possible to automatically generate and solve a generalized planning problem with subtasks. Work on the automatic generation of planning hierarchies~\cite{hogg2008htn} may be a good starting point. Another opportunity for future research is the identification of test cases that produce effective generalized plans for a given domain. This is also an open problem for FSMs~\cite{bonet2015policies} and, in general, for planning and learning systems~\cite{fern2011first,jimenez2012review}.


\begin{small}
\subsubsection*{Acknowledgments}
Sergio Jim\'enez is partially supported by the {\it Juan de la Cierva} program funded by the Spanish government.
\end{small}

\bibliographystyle{aaai}
\bibliography{paper-icaps16}

\end{document}